\documentclass[runningheads]{llncs}

\usepackage{eccv}
\usepackage{eccvabbrv}

\usepackage{graphicx}
\usepackage{booktabs}
\usepackage{bbding}
\usepackage[accsupp]{axessibility}  
\usepackage{cuted}
\usepackage{booktabs}
\usepackage{makecell}
\usepackage[pagebackref,breaklinks,colorlinks,citecolor=eccvblue]{hyperref}

\usepackage{orcidlink}

\begin{document}

\title{ScrollScape: Unlocking 32K Image Generation With Video Diffusion Priors
} 

\titlerunning{ScrollScape}

\author{Haodong Yu\inst{1} \and
Yabo Zhang\inst{1} \and
Donglin Di\inst{2} \and
Ruyi Zhang\inst{1} \and \\
Wangmeng Zuo\inst{1}$^{(}$\Envelope$^)$}
\authorrunning{H. Yu et al.}

\institute{Harbin Institute of Technology \and
Li Auto, Beijing, China\\
}
\maketitle
\begin{center}
    \centering
    \includegraphics[width=1.0\linewidth]{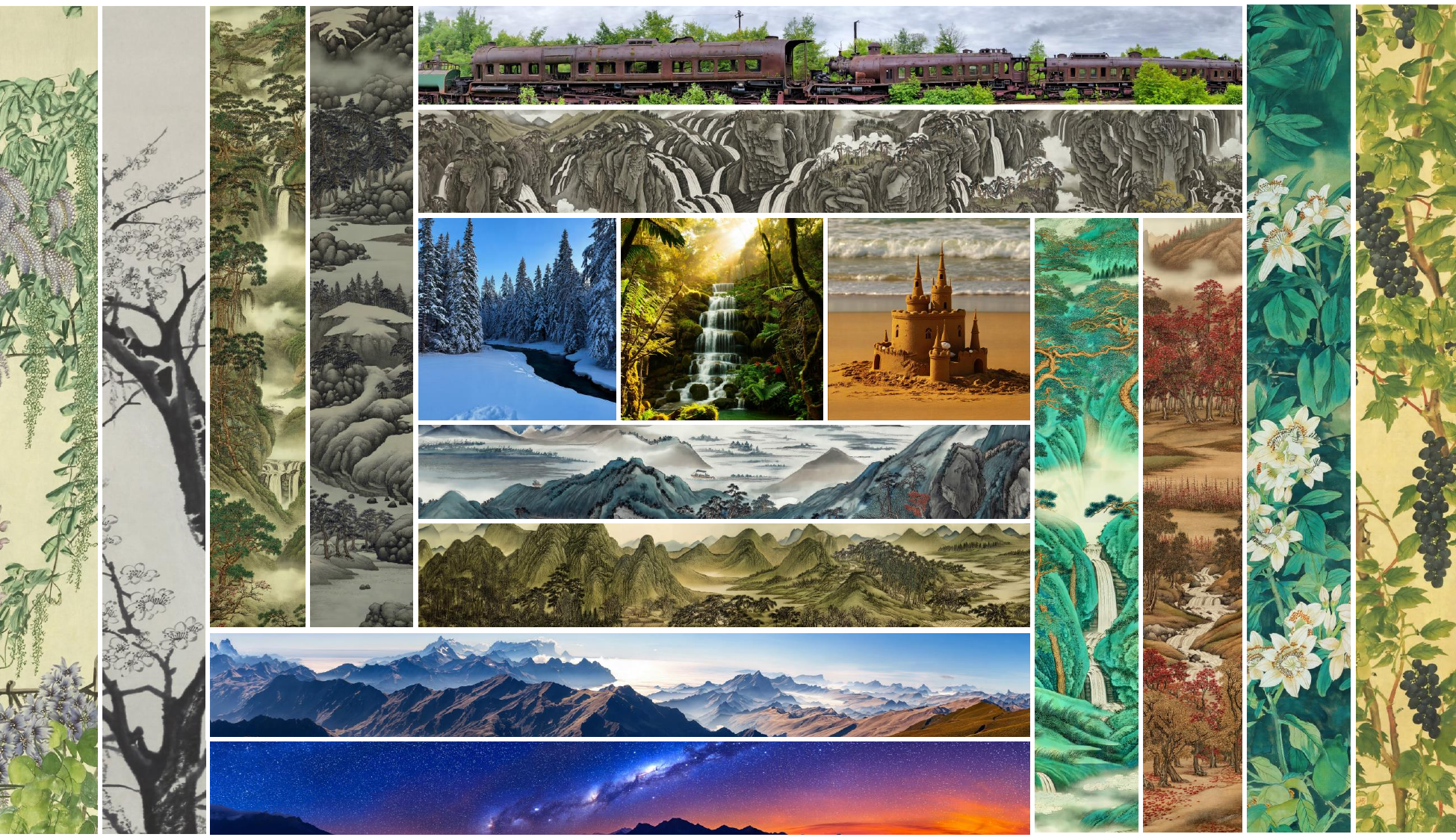}
    \captionof{figure}{\textbf{Imagery of ScrollScape.}  
  ScrollScape reformulates high resolution synthesis at extreme aspect ratios such as $8:1$ as a sequential video panning task. Leveraging robust video diffusion priors, it achieves exceptional $32\text{K}$ resolution across canvases ranging from traditional scrolls to photorealistic panoramas.}
    \label{fig:teaser}
\end{center}

\begin{abstract}
While diffusion models excel at generating images with conventional dimensions, pushing them to synthesize ultra-high-resolution imagery at extreme aspect ratios (EAR) often triggers catastrophic structural failures, such as object repetition and spatial fragmentation.
This limitation fundamentally stems from a lack of robust spatial priors, as static text-to-image models are primarily trained on image distributions with conventional dimensions.
To overcome this bottleneck, we present ScrollScape, a novel framework that reformulates EAR image synthesis into a continuous video generation process through two core innovations.
By mapping the spatial expansion of a massive canvas to the temporal evolution of video frames, ScrollScape leverages the inherent temporal consistency of video models as a powerful global constraint to ensure long-range structural integrity.
Specifically, Scanning Positional Encoding (ScanPE) distributes global coordinates across frames to act as a flexible moving camera, while Scrolling Super-Resolution (ScrollSR) leverages video super-resolution priors to circumvent memory bottlenecks, efficiently scaling outputs to an unprecedented 32K resolution.
Fine-tuned on a curated 3K multi-ratio image dataset, ScrollScape effectively aligns pre-trained video priors with the EAR generation task. 
Extensive evaluations demonstrate that it significantly outperforms existing image-diffusion baselines by eliminating severe localized artifacts. 
Consequently, our method overcomes inherent structural bottlenecks to ensure exceptional global coherence and visual fidelity across diverse domains at extreme scales.
\end{abstract}
\section{Introduction}
\label{sec:intro}
Diffusion models\cite{wan2025wan,zheng2024open,brooks2024video,yang2024cogvideox,blattmann2023stable,ho2022video,henschel2025streamingt2v,guo2023animatediff,kong2024hunyuanvideo,zhang2305controlvideo,wang2023videocomposer,sohl2015deep,song2021score,song2021denoising,ho2022classifier,ho2020ddpm,podell2023sdxl,lipman2022flow} have achieved remarkable success in synthesizing imagery with high quality and photorealistic detail. 
However, even the most preeminent models, such as Nano Banana Pro\cite{team2023gemini} and GPT-Image\cite{openai_gptimage_2025}, frequently suffer from catastrophic structural failure and fidelity loss when the target output deviates from standard dimensions. 
This limitation is particularly evident when these models are pushed to generate ultra-high-resolution imagery with extreme aspect ratios (EAR), such as landscape panoramas or traditional long scroll paintings.
We attribute this bottleneck to the fact that these foundational models are predominantly optimized on datasets with modest resolutions and conventional proportions. 
Consequently, they lack the robust spatial priors and long-range dependency modeling necessary to maintain global coherence across the massive, non-standard canvases required for such tasks.

Existing methods\cite{madar2025tiled,issachar2025dype,he2023scalecrafter,bar2023multidiffusion,haji2024elasticdiffusion,balaji2022ediffi,podell2023sdxl,bfl2024flux} attempt to modify the inference process or architecture of text-to-image diffusion models to synthesize EAR imagery at ultra high resolution. 
For the generation of extreme aspect ratios (\textit{e.g.}, $8:1$), Sync Diffusion\cite{lee2023syncdiffusion} partitions the target space into overlapping patches, which are subsequently processed piece-by-piece by the base model. 
However, this localized generation paradigm inherently struggles to guarantee global structural coherence, often leading to fragmented compositions. 
Furthermore, to achieve higher spatial resolutions, techniques like ScaleCrafter\cite{he2023scalecrafter} and DyPE\cite{issachar2025dype} manipulate the internal representations by dilating convolution modules and interpolating position embeddings. 
Although these adjustments expand the capacity for large-scale synthesis, the generation process remains highly unstable, frequently introducing severe artifacts such as object repetition. 
In essence, these training-free modifications demonstrate that relying solely on spatial priors derived from conventional image distributions is fundamentally insufficient to guarantee structural coherence at either ultra-high resolutions or extreme aspect ratios.

To break free from the inherent bottlenecks of purely image diffusion priors, we reformulate ultra-high-resolution EAR synthesis from a static image generation problem into a temporal video generation task. 
The core intuition is straightforward: we conceptualize the generation of a massive image as a continuous scanning process, translating the spatial layout of the canvas into a sequence of video frames. 
This paradigm shift enables us to harness the robust spatiotemporal priors of pre-trained video diffusion models\cite{wan2025wan,zheng2024open,yang2024cogvideox,zhang2305controlvideo,wang2023videocomposer,kong2024hunyuanvideo,bar2024lumiere,blattmann2023stable,ho2022video}. 
Unlike static text-to-image approaches that inevitably suffer from localized fragmentation and object repetition, the inherent temporal consistency of video models acts as a powerful global constraint, naturally ensuring long-range structural integrity and narrative continuity across massive canvases.

To seamlessly adapt video diffusion priors for EAR synthesis, we introduce ScrollScape, powered by Scanning Positional Encoding (ScanPE) and Scrolling Super-Resolution (ScrollSR).
First, standard video models assign identical spatial coordinates to every temporal frame. This rigid configuration anchors the generation process to a stationary viewpoint, rendering the model inherently incapable of panning across a massive canvas.
ScanPE overcomes this by distributing the global spatial coordinates of the target image across successive video frames. 
This re-engineering allows ScrollScape to act as a moving camera, synthesizing arbitrary aspect ratios through flexible scanning strategies.
Second, to tackle the immense computational cost of ultra-high-resolution generation, we propose ScrollSR. 
Rather than directly synthesizing massive pixels, ScrollSR leverages video super-resolution diffusion priors\cite{zhuang2025flashvsr,zhou2024upscale} to refine the content frame-by-frame. 
This efficient approach circumvents memory bottlenecks, successfully scaling the final output to an unprecedented $32\text{K}$. 
Finally, to train ScrollScape for EAR generation, we curated a lightweight dataset of approximately $3,000$ high-resolution, multi-ratio images. Empirical evidence confirms that this highly curated data is sufficient to perfectly align the pre-trained video model with the EAR synthesis objective.

Extensive evaluations demonstrate that ScrollScape achieves unprecedented high-fidelity EAR synthesis at up to $32\text{K}$ resolution, significantly outperforming existing approaches based on image diffusion models. 
Most notably, our method fundamentally overcomes the severe artifacts plaguing these baselines, including localized content repetition and structural fragmentation. 
Consequently, it ensures exceptional global coherence across diverse visual domains, ranging from photorealistic landscapes to stylized artistic scrolls.

To summarize, our key contributions are three-fold:
\begin{itemize}
    \item We introduce ScrollScape, a novel framework that reformulates EAR image synthesis as a sequential video panning process, effectively harnessing the video diffusion priors to ensure long-range structural coherence.
    \item We propose Scanning Positional Encoding (ScanPE) to re-engineer coordinate distributions for flexible, continuous generation across arbitrary aspect ratios, alongside a Scrolling Super-Resolution (ScrollSR) module that leverages video super-resolution diffusion priors frame-by-frame to achieve an unprecedented $32\text{K}$ resolution.
    \item Extensive experiments demonstrate that ScrollScape successfully eliminates severe artifacts like object repetition and fragmentation. Furthermore, we contribute a curated dataset of $3,000$ high-resolution, multi-ratio imagery to facilitate model adaptation and future research.
\end{itemize}
\begin{center}
  \centering
  \includegraphics[width=1.0\textwidth]{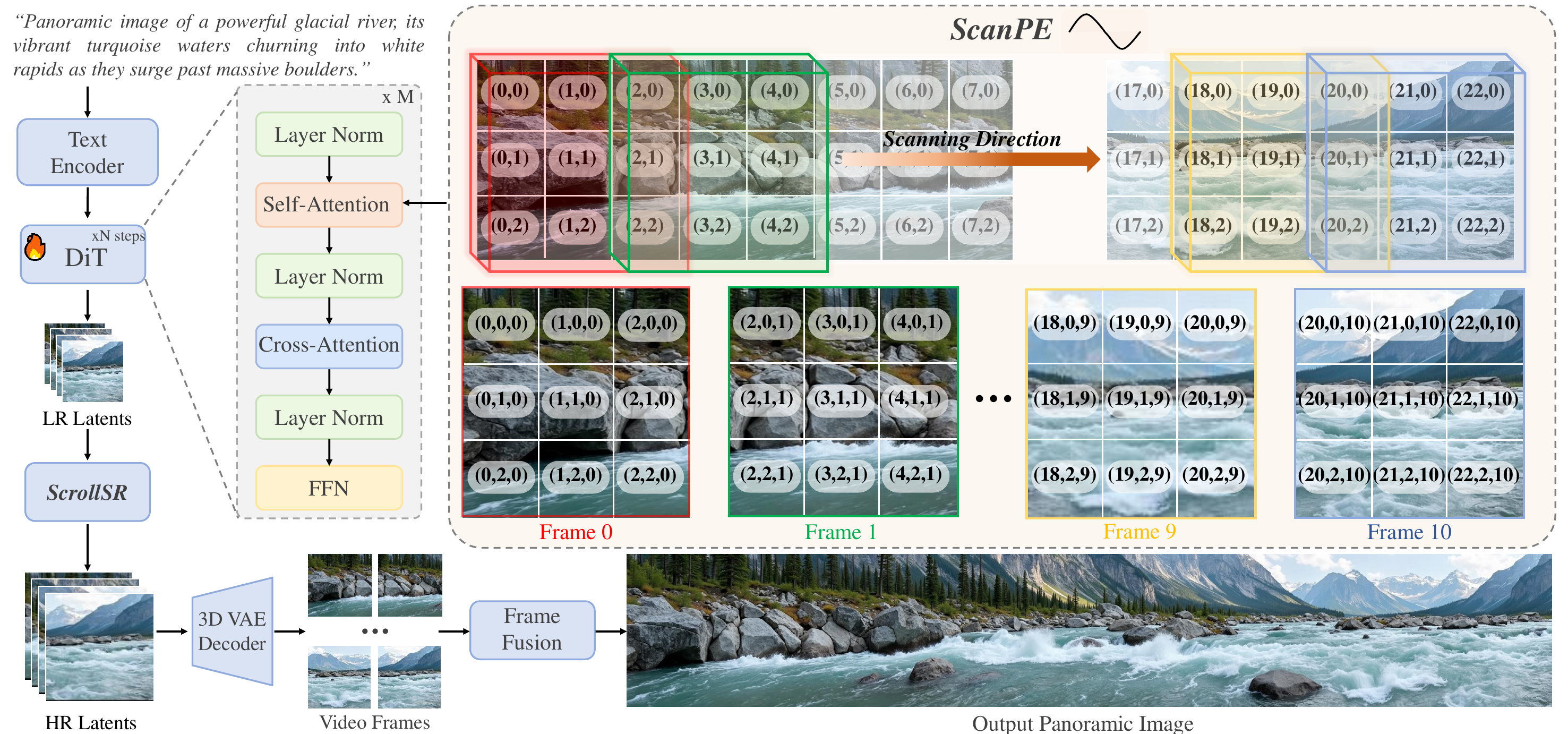}
  \captionof{figure}{\textbf{Overview of ScrollScape Framework.} 
 ScrollScape reformulates high resolution EAR synthesis as a sequential video panning task. \textbf{ScanPE} re engineers coordinate distributions by mapping global spatial indices $(x, y)$ onto a temporal sequence to ensure structural coherence across massive scales without the repetition typical of standard models.After generating low resolution latents via a hierarchical DiT, the \textbf{ScrollSR} module utilizes video super resolution priors to enhance details frame by frame. Finally, a 3D VAE decoder and frame fusion stage produce seamless, photorealistic panoramas and traditional scrolls at an exceptional $32\text{K}$ resolution.
  }
  \label{fig:pipeline_overview}
\end{center}
\section{Related Work}
\subsection{High Resolution EAR Synthesis}
The synthesis of high resolution imagery with extreme aspect ratios (\textit{e.g.}, $1:8$) remains a significant challenge for text-to-image models~\cite{zhang2017stackgan,esser2024scaling,balaji2022ediffi,kang2023gigagan,sauer2023stylegan,chang2023muse,yu2022scaling,nichol2022glide,ramesh2022hierarchical}.
For extreme aspect ratio generation, existing methods \cite{lee2023syncdiffusion,bar2023multidiffusion,madar2025tiled} partition the target space into overlapping patches. However, this localized generation paradigm inherently lacks a unified global structural prior, leading to the spatial fragmentation and disjointed compositions typical of window based methods.

Furthermore, to achieve higher spatial resolutions, techniques like ScaleCrafter \cite{he2023scalecrafter} and DyPE \cite{issachar2025dype} manipulate internal representations through dilated convolutions or positional interpolation.
While these adjustments expand the synthesis capacity, they fail to resolve the stationary bias of static architectures.
Consequently, the generation remains unstable, frequently triggering catastrophic structural failures such as object repetition when pushed beyond the model's original training distribution.
In essence, these modifications demonstrate that relying solely on stationary spatial priors of static text-to-image models is fundamentally insufficient to guarantee global structural coherence at extreme scales. 
This technological bottleneck necessitates our paradigm shift toward leveraging spatiotemporal video diffusion priors for high resolution EAR synthesis at an unprecedented 32K scale.

\subsection{Video Diffusion Models as Generative Priors}
Video diffusion models \cite{wan2025wan,zheng2024open,brooks2024video,yang2024cogvideox,blattmann2023stable,ho2022video,henschel2025streamingt2v,guo2023animatediff,kong2024hunyuanvideo,zhang2305controlvideo,wang2023videocomposer,ramesh2021zero,rombach2022high,saharia2022photorealistic} have demonstrated a significantly stronger capacity for simulating real world dynamics compared to image diffusion models.
While primarily designed for video generation, recent studies have begun to harness these robust priors for image centric applications, such as multi view synthesis \cite{voleti2024sv3d, li2024era3d} and image editing \cite{Frame2Frame,alzayer2025magic,zhang2025framepainter}.Specifically, Frame2Frame\cite{Frame2Frame}, Magic Fixup\cite{alzayer2025magic}, and Framepainter\cite{zhang2025framepainter} typify this trend by reformulating image editing as a temporal trajectory or a video-based propagation process. Frame2Frame\cite{Frame2Frame}, for instance, leverages a generative video model to create a continuous path from an initial state to a target state, thereby ensuring zero-shot coherence throughout the editing process.

However, existing explorations predominantly focus on leveraging temporal consistency for frame-to-frame transitions. 
The potential of these priors to regulate long distance spatial consistency as well as the utilization of video super-resolution priors for massive scaling remains largely underexplored.
ScrollScape diverges from these approaches by uniquely dual purposing video knowledge. 
We first utilize video diffusion priors to generate structurally coherent extreme aspect ratio (EAR) imagery, effectively transforming temporal continuity into a spatial regulator.
Building on this foundation, we are the first to leverage video super-resolution (VSR) diffusion priors to scale the results to an unprecedented 32K resolution.

\subsection{Positional Encoding in Transformers}
\label{Sec:RoPE}
The structural coherence of Diffusion Transformers (DiT) \cite{ma2024latte,hatamizadeh2024diffit,peebles2023scalable,vaswani2017attention} is fundamentally governed by their positional encoding\cite{shaw2018self} schemes.
In video DiTs, Rotary Positional Embedding (RoPE) \cite{su2024roformer,chen2023extending,peng2023yarn,ding2024longrope} has become the de facto standard due to its ability to model relative spatiotemporal dependencies.
By representing positions as rotations in the embedding space, RoPE ensures that attention weights depend strictly on relative displacement.
The efficacy of RoPE is highly susceptible to sequence length, particularly in high resolution EAR synthesis where inference scales often exceed the original training distribution. 

As positional indices shift into out of distribution phases, the model's global structural awareness is compromised. 
This distributional shift frequently manifests as object repetition and spatial fragmentation, mirroring the structural failures seen in static architectures when pushed beyond their design limits.
Standard 3D-RoPE is typically restricted to a static viewport, anchoring spatial indices to fixed frame centric grids.
This invariant mapping fails to guide the global spatial evolution required for expansive canvases, particularly within a sliding window paradigm that must track global trajectories while maintaining stable biases. 
To bridge this gap, we introduce ScanPE, a flexible coordinate system that re-engineers positional distributions to enable the moving camera vision for high resolution EAR synthesis.

\section{Method}
We propose ScrollScape, a framework that reformulates ultra-high-resolution EAR imagery synthesis  as a sequential generation task by leveraging video diffusion priors(in \cref{Sec:3.1}). 
The framework projects the spatial extension of long scroll imagery onto the temporal domain of a video sequence, allowing the model to utilize video diffusion priors for consistent sequential progression.
Specifically, ScrollScape employs Scanning Positional Encoding (in \cref{Sec:3.2}) to maintain spatial consistency throughout the sequential generation, yielding imagery with structural coherence and rich content, even at extreme aspect ratios.
Building upon this coherent foundation, ScrollSR (in \cref{Sec:3.3}) directly leverages Diffusion based video super-resolution priors without any additional training for resolution enhancement and detail augmentation.

\subsection{Ultra Image Generation as Video Scanning}\label{Sec:3.1}
High resolution EAR synthesis aims to generate content across extreme aspect ratios such as digital scrolls and panoramic landscapes. The fundamental challenge lies in maintaining global coherence across an exceptionally large spatial span. We observe that the spatial extension of such imagery can be effectively modeled as a temporal progression. Since video sequences naturally capture the structural continuity of the real world, video diffusion priors provide the ideal foundation for maintaining consistency across extended canvases. 

To this end, we reformulate EAR synthesis as a sequential generation task by projecting spatial coordinates onto the temporal domain.Unlike traditional text-to-image frameworks, ScrollScape leverages video diffusion priors to model spatial extension as a continuous temporal progression. Specifically, given a pretrained VAE, the synthesis is performed in the latent space $\mathcal{Z}$. The global panoramic latent $\mathbf{z} \in \mathbb{R}^{h \times w \times c}$ is partitioned into a sequence of $N$ overlapping chunks $\mathcal{S} = \{\mathbf{z}_t\}_{t=1}^N$, where each chunk $\mathbf{z}_t \in \mathbb{R}^{h \times l \times c}$ is defined by a fixed window length $l$ and is treated as the $t$-th temporal state of the video diffusion prior. To establish a unified spatial structure, each chunk is anchored to a global horizontal coordinate $\phi_t$ via the mapping:
\begin{equation}
\phi_t = (t-1) \cdot \Delta, \quad t \in \{1, \dots, N\},
\end{equation}
where $\Delta$ denotes the spatial stride. Consequently, the $t$-th window covers the spatial interval $\Omega_t = [\phi_t, \phi_t + l]$, ensuring that the spatial progression of the imagery is semantically aligned with the temporal dynamics of the model. By strictly binding spatial displacement to temporal indices, this formulation effectively bypasses the stationary bias of static architectures, transforming what would otherwise be a repetitive tiling process into a globally coherent narrative flow across an expansive digital canvas.
 
\subsection{Scanning Positional Encoding}
\label{Sec:3.2}
Although sequential video panning enables high resolution EAR synthesis, standard 3D RoPE lacks the precise global coordinate control required to track spatial evolution across successive frames. 
This deficiency forces the model to operate within a frame centric coordinate system, limiting its ability to maintain long range structural integrity across massive canvases. 
To resolve this, we introduce Scanning Positional Encoding (ScanPE), which generalizes the static position $\mathbf{p}$ into a dynamic scanning trajectory. 
By mapping tokens into a unified global coordinate system through spatial displacements, ScanPE re engineers the model to act as a moving camera.
To characterize the sequential progression of the scanning trajectory, we define $t \in \{1, \dots, N\}$ as the temporal block index. The global anchor position $\mathbf{O}_t$ is formulated as a discrete coordinate accumulation:
\begin{equation}
    \mathbf{O}_t = \sum_{k=1}^{t-1} \delta \cdot \mathbf{d}_k + \mathbf{P}_{init}, \quad \text{with } \mathbf{O}_1 = \mathbf{P}_{init},
\end{equation}
where $\delta$ represents the step wise stride and $\mathbf{d}_k \in \{ (\pm 1, 0), (0, \pm 1) \}$ denotes the unit direction vector. This formulation unifies diverse scanning behaviors: in Linear Mode, $\mathbf{d}_k$ remains constant to follow a unidirectional path, while in Snake Fusion Mode, $\mathbf{d}_k$ follows a periodic switching function to cover complex manifolds on non standard canvases.To implement the moving camera paradigm, we project frame centric local coordinates $\mathbf{p}_{loc} = (h_{loc}, w_{loc})$ into a unified global coordinate system $\mathbf{P}_g = (H_g, W_g)$:
\begin{equation}
\mathbf{P}_g(t, \mathbf{p}_{loc}) = \mathbf{p}_{loc} + \mathbf{O}_t.
\end{equation}
The final rotational embedding $\mathcal{R}$ is synthesized by coupling these trajectory aware coordinates with the frequency basis $\theta_j = \beta^{-2j/d}$ where $\beta$ denotes the base frequency constant. By substituting these phases into the rotation operator $\mathbf{\Theta}$, we obtain the spatiotemporally decoupled embedding:
\begin{equation}
    \mathcal{R}(t, H_g, W_g) = \text{Concat} \left[ \mathbf{\Theta}(t \theta_j), \mathbf{\Theta}(H_g \theta_j), \mathbf{\Theta}(W_g \theta_j) \right].
\end{equation}
While ScanPE addresses phase drift, directly applying video diffusion priors to EAR synthesis introduces a distributional mismatch that often leads to localized artifacts. To mitigate this, we perform a lightweight alignment stage using conditional Flow Matching.Given an EAR latent sample $z_0 \sim p_{\text{data}}$ and Gaussian noise $z_1 \sim \mathcal{N}(0,\mathbf{I})$, we define a linear interpolation path $z_\tau = (1-\tau) z_0 + \tau z_1$.We optimize a conditional vector field $v_\theta$ via:
\begin{equation}
\mathcal{L}_{\text{FM}} =\mathbb{E}_{z_0, z_1, \tau}
\left[\left\|v_\theta(z_\tau, \tau, c, \mathcal{R})-(z_1 - z_0)\right\|_2^2\right],
\end{equation}
where $c$ denotes the text prompt and $\mathcal{R}$ represents the trajectory aware embeddings.This alignment stage fine tunes the model to handle extreme aspect ratio distributions while regularizing the generative process with its pre trained spatiotemporal knowledge. 
\subsection{Scrolling Super Resolution}\label{Sec:3.3} 
To elevate visual fidelity, we introduce Scrolling Super Resolution (ScrollSR), which performs high resolution refinement directly within the latent space $\mathcal{Z}$. Rather than upscaling synthesized pixels, ScrollSR leverages video super-resolution diffusion priors to transform low resolution latents into high resolution counterparts. This latent space operation is critical for $32\text{K}$ synthesis, as it refines intricate details while circumventing the immense memory overhead associated with high resolution pixel space optimization.While latent based refinement ensures computational efficiency, the subsequent translation into the pixel domain is hindered by the 3D VAE, which tends to introduce flickering or boundary artifacts when decoding adjacent frames with significant pixel variance. 
To eliminate this influence, we employ Trajectory Anchored Partitioning (TAP) as a zero shot spatial alignment strategy. 

Instead of processing the latent sequence as a uniform stream, TAP serves as a specialized partitioning scheme that maps windows corresponding to unique global spatial coordinates onto discrete latent blocks.This strategy ensures that a Primary Anchor $\mathbf{z}_1$ stabilizes the initial decoder state while subsequent blocks are anchored to their respective spatial indices established by ScanPE. By regularizing the latent trajectory across the sequence through this spatial grouping, TAP achieves seamless synchronization between the 3D VAE decoder and our coordinate system without requiring additional training. Consequently, ScrollSR produces high fidelity output that is free from localized artifacts and geometric distortions while preserving strict global structural alignment.

\subsection{Panoramic Frame Fusion}
Following the latent-to-pixel reconstruction guided by TAP, we integrate the generated sequential blocks into a unified 32K panorama via a weighted fusion strategy. This stage resolves residual inconsistencies across overlapping sliding windows while suppressing flickering and transient disturbances introduced during sequential decoding.

For each temporal block $t$ containing $N$ frames, we employ Median Consensus Selection (MCS) to extract a structurally stable representative tile $\bar{I}_t$. Unlike temporal averaging, which may accumulate boundary noise or scanning-induced motion blur, MCS selects the frame whose feature statistic is closest to the temporal median:
\begin{equation}
\bar{I}_t
=
\arg\min_{i}
\left|
f_{t,i}
-
\operatorname{Median}\{ f_{t,k} \}_{k=1}^N
\right|.
\end{equation}
Here $f_{t,i}$ denotes the $i$-th frame within block $t$. This selection strategy favors the frame exhibiting maximal temporal consensus, thereby filtering out outliers while preserving high-frequency structural details.

The final panorama $\mathcal{I}_{\text{pano}}$ is reconstructed by projecting each distilled tile $\bar{I}_t$ onto the global coordinate system:
\begin{equation}
\mathcal{I}_{\text{pano}}(\mathbf{P}_g)
=
\frac{\sum_t M(\mathbf{P}_g - \mathbf{O}_t)\,\bar{I}_t(\mathbf{P}_g - \mathbf{O}_t)}
{\sum_t M(\mathbf{P}_g - \mathbf{O}_t)},
\end{equation}
where $\mathbf{P}_g$ denotes the global pixel coordinate. The weight mask $M$ adopts a distance-based ramp to ensure smooth transitions between adjacent windows.

\begin{center}
    \centering
    \includegraphics[width=1.0\linewidth]{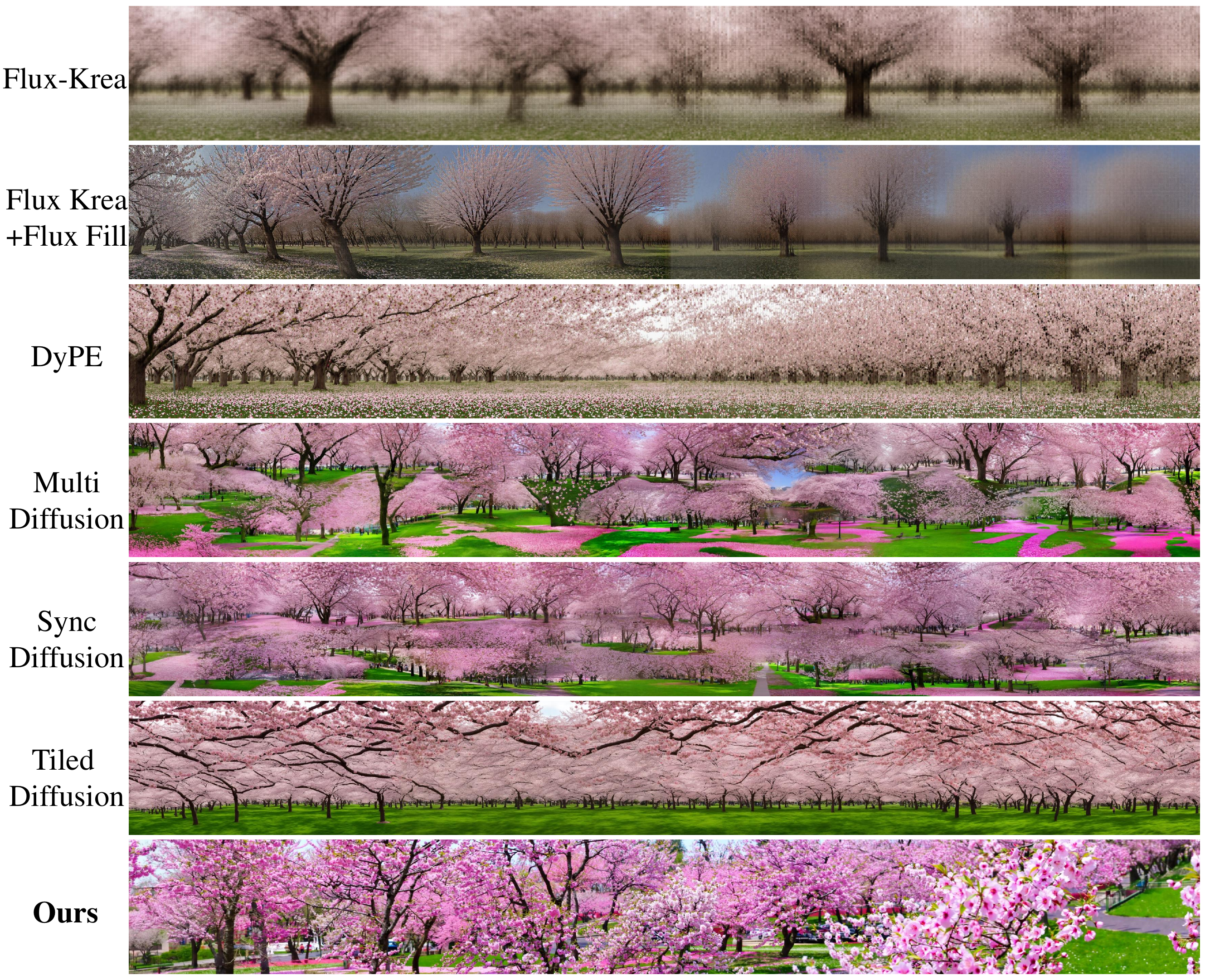}
    \captionof{figure}{\textbf{Qualitative comparison on 8:1 horizontal panoramic scroll imagery.} 
    ScrollScape achieves rigorous structural coherence and expansive global diversity, whereas baselines suffer from the semantic repetition and boundary artifacts common in tiled synthesis.}
    \label{fig:8_1compare}
\end{center}

\section{Experiments}
\subsection{Experimental Setup}
\noindent\textbf{Dataset Construction.} 
To support the synthesis of EAR content, we curate a specialized training dataset consisting of 3,000 high-quality panoramic samples. 
Specifically, this includes 2,000 natural landscape images with resolution ratios of 6:1 or higher and 1,000 traditional Chinese landscape paintings maintained at a consistent 6:1 ratio.
These samples provide the model with essential compositional priors for continuous, horizontal-flowing structures, ranging from geological vistas to traditional ink-wash scroll aesthetics.
\begin{figure}[!t]
    \centering
    \includegraphics[width=0.95\linewidth]{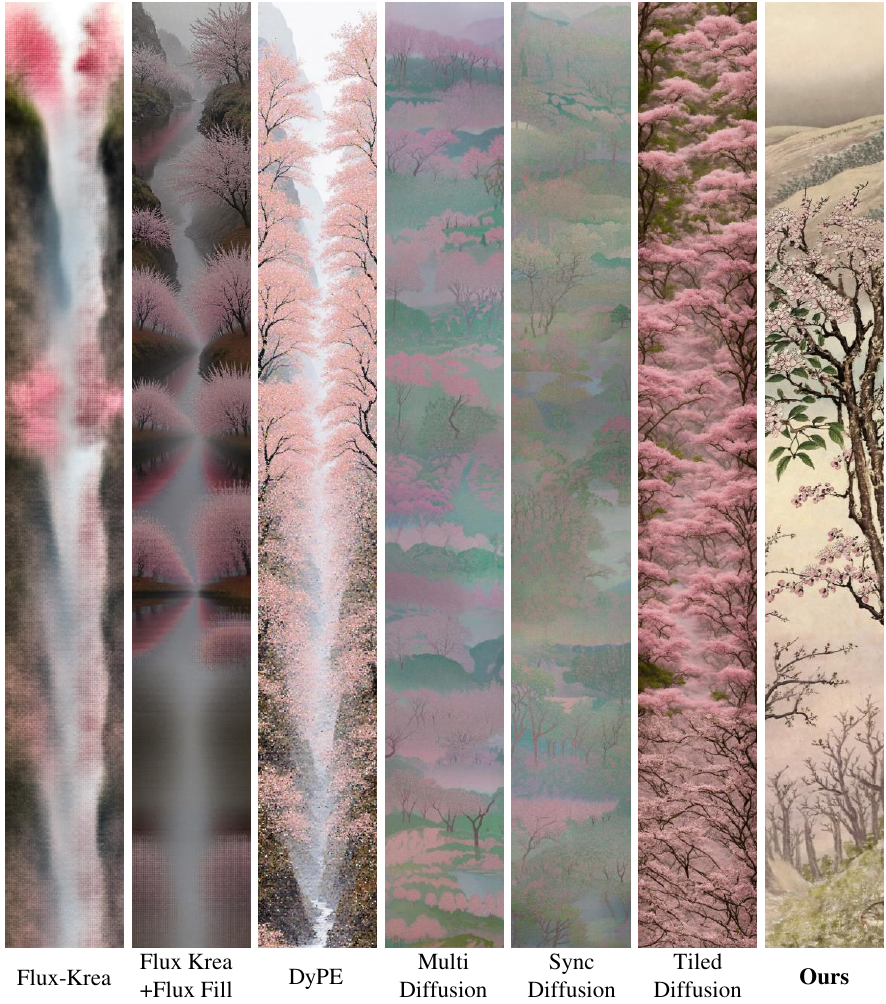}
    \captionof{figure}{\textbf{Qualitative results on 8:1 vertical scroll paintings.}
    ScrollScape maintains superior global structural coherence and expansive diversity, effectively eliminating the boundary artifacts and semantic repetition found in existing frameworks.}
    \label{fig:8_1compare_vertical}
\end{figure}
\begin{figure}[!t]
  \centering
  \includegraphics[width=1.0\linewidth]{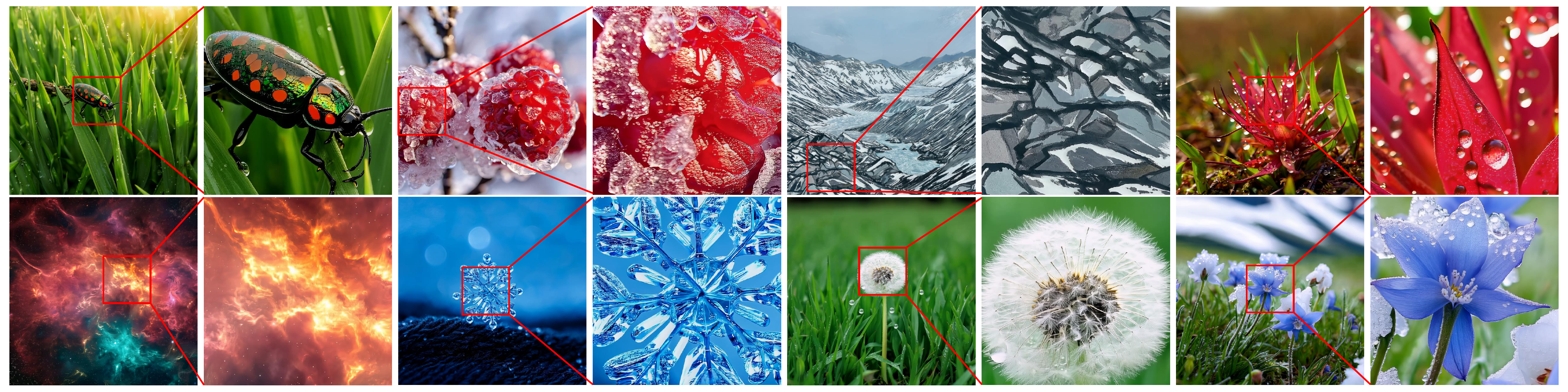}
  \captionof{figure}{\textbf{Demonstration of high fidelity 8K imagery generated by ScrollScape.}
  The results showcase the versatility of \textbf{ScrollScape} in producing high fidelity imagery with structural richness and local clarity across various subject matters, ranging from microscopic textures to macroscopic landscapes.\textit{Please zoom in for high resolution details.}}
  \label{fig:8Kresults}
\end{figure}

\noindent\textbf{Implementation Details.}We initialize ScrollScape with Wan2.1-T2V-1.3B\cite{wan2025wan} and finetune it on our collected training samples. We train the model on two A100 GPUs with a total batch size of 4. The model is optimized for 10, 000 iterations with a learning rate of $1 \times 10^{-5}$ using the AdamW\cite{loshchilov2017decoupled} algorithm. During inference, the model generates base pixel frames at a reduced resolution to maintain global coherence. These frames are subsequently processed by the modified FlashVSR\cite{zhuang2025flashvsr} module on a single A100 (80GB) GPU to reach the final 32K resolution. We use a linear ramp function for the pixel-level fusion of overlapping tiles to ensure a seamless transition across the expansive panorama.

\subsection{Comparison with Baselines}
\noindent \textbf{Qualitative Results.}
\Cref{fig:8_1compare,fig:8_1compare_vertical} present a qualitative comparison of panoramas synthesized at an extreme aspect ratio of 8:1, with the short side resolution maintained at 1024 pixels.
While existing methods struggle to maintain quality under such constraints, ScrollScape demonstrates a significant leap in both structural integrity and content richness. Specifically, diffusion based approaches such as DyPE\cite{issachar2025dype}, FLUX.1-Krea-dev\cite{bf2024fluxkrea}, and FLUX.1-Fill-dev\cite{bfl2024fluxfill} exhibit severe visual artifacts and geometric distortions once the aspect ratio exceeds 4:1. Meanwhile, tiling based frameworks, including MultiDiffusion\cite{bar2023multidiffusion}, SyncDiffusion\cite{lee2023syncdiffusion}, and Tiled Diffusion\cite{madar2025tiled}, suffer from conspicuous content repetition and irrational global structures, failing to establish a coherent narrative flow. In contrast, ScrollScape maintains robust structural coherence and diverse, non redundant details across various themes. This performance demonstrates a superior balance between local fidelity and global architectural logic, significantly outperforming competitive baselines.

As shown in \cref{fig:8Kresults}, ScrollScape maintains global structural integrity and high fidelity details at 8K resolution. Beyond panoramic synthesis, the hybrid mode ensures standard ratio generation with consistent structural layout and clarity. Zoomed in examples, such as beetle carapaces and ice crystals, demonstrate sharp symmetry and intricate textures, proving that ScrollScape preserves global architectural logic alongside fine grained details within a single unified architecture.

\noindent \textbf{Quantitative Results.}
As demonstrated in \cref{tab:benchmark8_1}, we adopt a patch based assessment strategy to mitigate geometric distortion and aspect ratio bias during evaluation. Specifically, we partition the $8:1$ panoramas into non-overlapping spatial grids with a 1:1 aspect ratio. Local fidelity and distribution similarity are quantified by averaging FID \cite{heusel2017gans} and KID \cite{binkowski2018demystifying} scores across these patches. Specifically, both metrics are computed by comparing the patches against images from Aesthetic Eval\cite{zhang2025diffusion} after resizing them to match the corresponding dimensions. Semantic alignment is evaluated via CLIP scores \cite{hessel2021clipscore}. To ensure seamless junctions, we assess stylistic uniformity and visual continuity using Intra Style Loss\cite{gatys2015neural} between adjacent grids.We introduce Global Structural Diversity (GSD) to quantify long range architectural variety and detect mode repetition across the panoramic manifold. Rather than focusing on local fidelity, GSD evaluates the correlation between distant spatial grids via two complementary perspectives. Specifically, we utilize LPIPS \cite{zhang2018unreasonable} distances to capture perceptual texture variation and DINOv2 \cite{oquab2023dinov2} feature cosine similarity to assess semantic non redundancy. By identifying high feature similarity between spatially separated regions, GSD effectively exposes the semantic looping and object duplication inherent in tiling based frameworks. Ultimately, this metric confirms that ScrollScape preserves a consistent artistic style while continuously synthesizing unique and varied structural elements throughout the entire expansive sequence.The experimental results show that ScrollScape maintains high quality details while effectively avoiding the content repetition typical of traditional generative methods.
\begin{table}[!t]
    \centering
    \caption{\textbf{Quantitative comparison.} ScrollScape surpasses alternative approaches across all metrics. The best results are \textbf{bolded}.}
    \label{tab:benchmark8_1}
    \footnotesize 
    \renewcommand{\arraystretch}{1.0} 
    \setlength{\tabcolsep}{3pt}
    \begin{tabular}{@{} l cccccc @{}} 
        \toprule
        Method & FID $\downarrow$ & CLIP $\uparrow$ & \makecell[c]{KID $\downarrow$ \\ \scriptsize$(\times 10^{-2})$} & \makecell[c]{Style-L $\downarrow$ \\ \scriptsize$(\times 10^{-3})$} & \makecell[c]{GSD $\uparrow$ \\ \scriptsize(LPIPS)} & \makecell[c]{GSD $\downarrow$ \\ \scriptsize(DINOv2)} \\
        \midrule
        FLUX-Krea       & 333.7 & 20.9 & 20.4 & 6.0 & 0.174 & 0.822 \\
        FLUX-Krea+Fill  & 281.4 & 19.1 & 4.9 & 13.5 & 0.293 & 0.841 \\
        DyPE            & 248.1 & 24.6 & 4.7 & 5.5 & 0.569 & 0.682 \\
        MultiDiffusion  & 261.7 & 29.7 & 3.7 & 5.0 & 0.658 & 0.902 \\
        SyncDiffusion   & 245.2 & 26.5 & 3.2 & 4.7 & 0.618 & 0.895 \\
        Tiled Diffusion & 241.2 & 27.3 & 3.0 & 4.5 & 0.591 & 0.901 \\
        \midrule
        \textbf{ScrollScape (Ours)} & \textbf{214.7} & \textbf{30.0} & \textbf{2.0} & \textbf{4.0} & \textbf{0.674} & \textbf{0.670} \\
        \bottomrule
    \end{tabular}
\end{table}

\begin{table}[!t]
  \caption{\textbf{User Preference Study.}
  The percentages denote the fraction of raters who prefer our results over each baseline.}
  \label{tab:main_user}
  \centering
  \begin{tabular}{@{}lccc@{}}
    \toprule
    Method Comparison & Structural Coh. & Content Rich. & Image Qual. \\
    \midrule
    Ours vs DyPE & 92\% & 89\% & 87\% \\
    Ours vs MultiDiffusion & 83\% & 76\% & 82\% \\
    Ours vs SyncDiffusion & 85\% & 74\% & 78\% \\
    Ours vs Tiled Diffusion & 76\% & 85\% & 74\% \\
    \bottomrule
  \end{tabular}
\end{table}

\begin{table}[!t]
    \centering
    \caption{\textbf{Quantitative ablation study.} While Wan2.1 achieves an outstanding Style-L score due to mode repetition, ScrollScape demonstrates superior performance across most metrics, ensuring both high fidelity and global structural diversity.}
    \label{tab:ablation}
    \footnotesize 
    \setlength{\tabcolsep}{3.8pt} 
    \begin{tabular}{@{} l cccccc @{}} 
        \toprule
        Method & FID $\downarrow$ & CLIP $\uparrow$ & \makecell[c]{KID $\downarrow$ \\ \scriptsize$(\times 10^{-2})$} & \makecell[c]{Style-L $\downarrow$ \\ \scriptsize$(\times 10^{-3})$} & \makecell[c]{GSD $\uparrow$ \\ \scriptsize(LPIPS)} & \makecell[c]{GSD $\downarrow$ \\ \scriptsize(DINOv2)} \\
        \midrule
        Wan2.1          & 246.5 & 25.9 & 4.7 & \textbf{2.0} & 0.303 & 0.975 \\
        w/o Training      & 272.0 & 24.6 & 6.3 & 7.6 & 0.569 & 0.858 \\
        w/o TAP           & 231.6 & 25.6 & 3.2 & 6.0 & 0.469 & 0.730 \\
        w/o ScrollSR      & 218.8 & 26.5 & 2.4 & 4.9 & 0.668 & 0.671 \\
        \midrule
        \textbf{ScrollScape (Ours)} & \textbf{214.7} & \textbf{30.0} & \textbf{2.0} & 4.0 & \textbf{0.674} & \textbf{0.670} \\
        \bottomrule
    \end{tabular}
\end{table}
\begin{figure}[!t]
  \centering
  \includegraphics[width=1.0\linewidth]{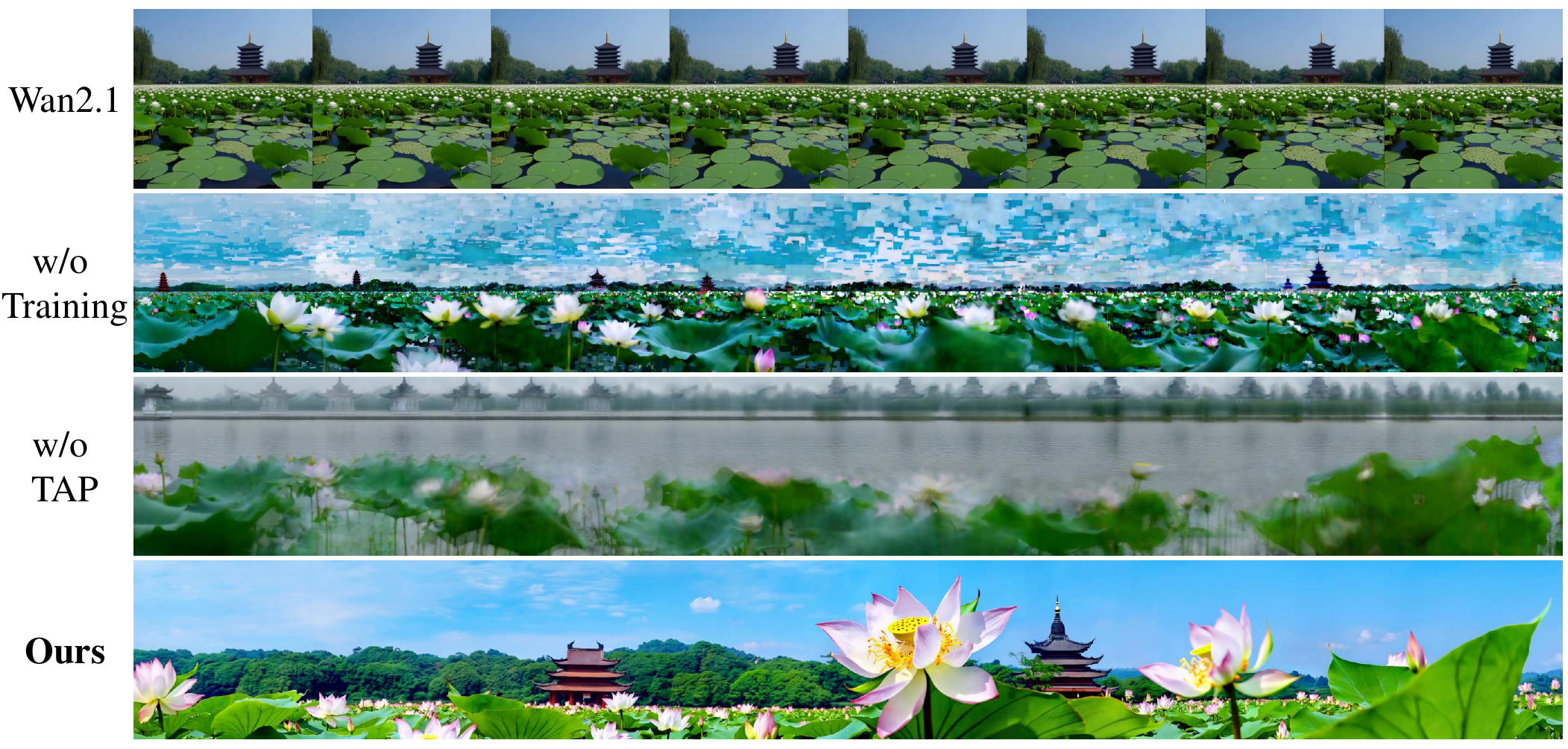}
  \captionof{figure}{\textbf{Qualitative ablation study on the effectiveness of ScanPE.}
  From top to bottom: (a) the vanilla Wan2.1 exhibiting severe content repetition; (b) w/o training, where ScanPE successfully breaks repetitive patterns but yields chaotic textures; (c) w/o TAP showing significant blurring due to decoding limitations; and (d) ScrollScape producing seamless panoramas with structural coherence and rich details.
  }
  \label{fig:ablation_qual}
\end{figure}
\begin{figure}[!t]
  \centering
  \includegraphics[width=1.0\linewidth]{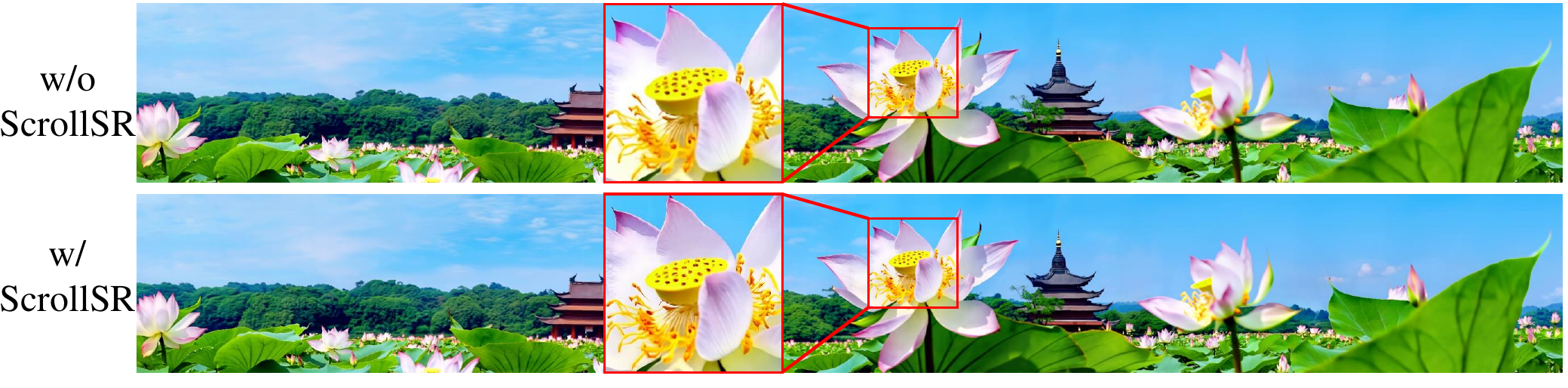}
  \captionof{figure}{\textbf{Qualitative ablation study on the effectiveness of ScrollSR.}The baseline w/o ScrollSR (top) suffers from perceptual blurriness in the lotus petals. In contrast, our full framework w/ ScrollSR (bottom) successfully restores fine grained textures and structural definition, as evidenced by the sharp details and high resolution clarity in the zoomed in patches.
  }
  \label{fig:ablation_SR}
\end{figure}

\noindent \textbf{User Study.}
To evaluate the perceptual quality of ScrollScape, we conducted a user study utilizing $100$ diverse prompts spanning multiple themes. For each prompt, we generated panoramas with an 8:1 aspect ratio in both horizontal and vertical orientations to compare our method against existing baselines. Twenty independent raters performed anonymized side by side evaluations across three primary dimensions: (i) structural coherence for logical and seamless global structure, (ii) content richness for varied details without redundancy, and (iii) image quality for visual fidelity and texture clarity.
As summarized in \cref{tab:main_user}, ScrollScape is significantly preferred over baselines across all three perspectives.

\subsection{Ablation Study}
As illustrated in \cref{fig:ablation_qual}, the vanilla Wan2.1 model suffers from severe content repetition when generating sequences at extreme aspect ratios. Each segment appears as a near identical replica of the preceding one, failing to establish a meaningful spatial progression.The variant w/o Training underscores the vital role of ScanPE in maintaining structural diversity. By successfully breaking the repetitive patterns observed in the baseline, this variant proves that ScanPE provides a superior structural foundation for EAR synthesis. However, without the proposed training phase, the model produces chaotic textures and significant visual noise, demonstrating that training is essential to harmonize video diffusion priors with the high resolution static domain.

Similarly, the results of w/o TAP show significant blurring and structural collapse. This confirms that the standard decoder lacks the capacity to reconstruct fine grained details from the extended latent sequence produced during sequential generation. In contrast, our full model achieves exceptional structural coherence and content richness. As evidenced in \cref{fig:ablation_SR}, the w/o ScrollSR variant maintains global structural integrity but fails to achieve perceptual sharpness, which results in the visibly soft foreground textures. Our full model integrates ScanPE, TAP, and ScrollSR to resolve spatial and temporal ambiguities. This synergy yields non redundant panoramas with the high resolution clarity required for photorealistic synthesis.

\cref{tab:ablation} also highlights the effectiveness of each proposed module by providing a detailed numerical breakdown of their contributions. The quantitative results consistently align with our qualitative observations, reinforcing the necessity of our integrated approach for high-resolution EAR synthesis. For instance, consistent with previous observations, Wan's lower Style-L score is primarily due to the generation of a singular, repeated pattern.

\section{Conclusion}
In this paper, we reframe ultra-high-resolution EAR synthesis as a sequential video panning task and introduce ScrollScape to leverage the robust spatiotemporal priors of video diffusion models. Built upon Wan2.1-T2V-1.3B, ScrollScape effectively eliminates object repetition and spatial fragmentation while ensuring seamless global coherence. To address the limitations of static spatial indexing, we propose ScanPE to act as a moving camera and ScrollSR to efficiently scale outputs to an unprecedented $32\text{K}$ resolution. Our experiments highlight the effectiveness of this approach, which significantly outperforms existing image diffusion baselines across diverse domains. Furthermore, our framework demonstrates strong generalization by maintaining high fidelity details from microscopic textures to macroscopic landscapes. We hope our work will inspire further exploration of video priors as spatial regulators for high resolution generative tasks.

%
%
\bibliographystyle{splncs04}
\bibliography{main}
\end{document}